\documentclass[10pt, a4paper, twocolumn,showabstract]{naverlabseurope} % no full width teaser figure

\usepackage{lipsum}
\usepackage{multicol}
\usepackage{tikz,tkz-kiviat,pgfplots}

\usepackage{array, booktabs}
\usepackage{adjustbox}
\usepackage{longtable}
\usepackage{float}
\newcommand{\correct}[1]{\textcolor{blue}{ #1}}
\newcommand{\incorrect}[1]{\textcolor{red}{ #1}}

\graphicspath{{figures/}}

\title{\emph{StarDrinks}: An English and Korean Test Set for SLU Evaluation in a Drink Ordering Scenario}
\titlerunning{StarDrinks}

\correspondingauthor{stardrinks@naverlabs.com}

\authors{Marcely Zanon Boito, Caroline Brun, Inyoung Kim, Denys Proux, Salah Ait-Mokhtar, Nikolaos Lagos, Jean-Luc Meunier and Ioan Calapodescu}
\affiliations{NAVER LABS Europe}
\website{https://europe.naverlabs.com/stardrinks}
\websiteref{Dataset: }

\begin{abstract}
LLMs and speech assistants are increasingly used for task-oriented interactions, yet their evaluation often relies on controlled scenarios that fail to capture the variability and complexity of real user requests. Drink ordering, for example, involves diverse named entities, drink types, sizes, customizations, and brand-specific terminology, as well as spontaneous speech phenomena such as hesitations and self-corrections. To address this gap, we introduce \emph{StarDrinks}, a test set in English and Korean containing speech utterances features, transcriptions, and annotated slots. Our dataset supports speech-to-slots SLU, transcription-to-slots NLU, and speech-to-transcription ASR evaluation, providing a realistic benchmark for model robustness and generalization in a linguistically rich, real-world task.
\end{abstract}

\begin{document}

\maketitle

\section{Introduction}

LLMs and speech assistants have rapidly advanced in recent years, enabling increasingly natural and efficient interactions between humans and machines. These systems are now deployed in numerous everyday applications, including personal assistants, customer service bots, and automated ordering systems. Despite these impressive developments, the evaluation of such models is still largely conducted under controlled or simplified conditions that fail to reflect the diversity and unpredictability of real-world usage~\cite{lunardi2025robustness,pmlr-v235-wei24c,xiong2025stealtheval}. Standard benchmarks often rely on clean, well-structured inputs and limited vocabularies, which do not capture the full range of linguistic variability, noise, and contextual complexity encountered in spontaneous human speech.

One particular deployment setting where these limitations become particularly evident is drink ordering, which, while seemingly simple, involves complex linguistic phenomena and decision-making structures. Understanding and processing such requests requires a natural language understanding~(NLU) model to interpret the semantic intent behind user utterances, and in the case of spoken language understanding~(SLU), we add to this challenge the handling of speech input, including hesitations, self-corrections, disfluencies, and prosodic cues. Real drink orders often include multiple attributes such as drink type, size, temperature, milk preference, flavorings, and toppings, expressed in varying orders or using brand-specific terminology, further complicating semantic parsing.

Despite the importance of NLU in task-oriented dialogue systems, there is a notable shortage of datasets that reflect realistic spoken interactions, and, to the best of our knowledge, no publicly available dataset exists for spoken drink ordering. As a result, current models are often trained and evaluated on limited or artificial benchmarks, performing well in controlled settings but struggling to generalize to authentic, complex user interactions.

To address the gap in realistic evaluation resources, we release \emph{StarDrinks}, a test set in English and Korean for the drink-ordering scenario. \emph{StarDrinks} contains speech utterance features, corresponding transcriptions, and annotated slots, making it suitable for multiple evaluation settings. It can be used for speech-to-slots SLU evaluation of assistants, transcription-to-slots NLU evaluation, and speech-to-transcription ASR evaluation in a scenario that requires generalization to previously unseen named entities. The dataset is collected from authentic drink orders, capturing diverse named entities, complex combinations of order attributes, and natural speech phenomena such as hesitations, self-corrections, and disfluencies. By providing this rich and challenging test set, \emph{StarDrinks} enables more realistic assessments of LLMs and spoken language assistants, supporting the development of more robust, context-aware systems capable of handling real-world interactions.

This paper is organized as follows. Section~\ref{sec:relatedworks} reviews existing datasets for NLU and SLU. Section~\ref{sec:datacreation} describes the dataset creation process in detail. The resulting dataset is presented in Section~\ref{sec:dataset}, followed by a use case involving a drink-ordering agent in Section~\ref{sec:usecase}. Finally, Section~\ref{sec:conclusion} concludes the paper.

\begin{figure*}
\begin{center}
\includegraphics[width=\textwidth]{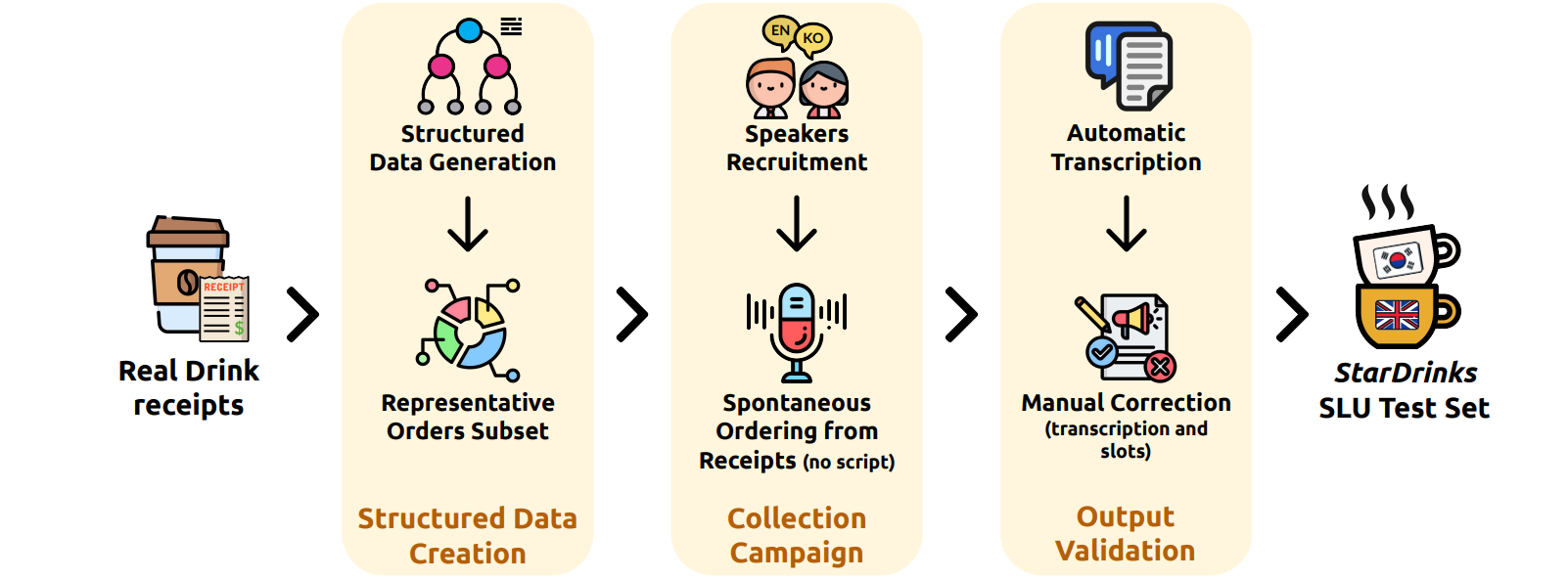}
\caption{\emph{StarDrinks} data generation pipeline overview.}
\label{fig:datapipeline}
\end{center}
\end{figure*}

\section{Related Works}\label{sec:relatedworks}

SLU and NLU datasets typically consist of speech and text data, respectively, accompanied by annotations for intent classification and/or slot filling tasks. Moreover, datasets can be either \textit{single} or \textit{multi-turn}, meaning that the task is accomplished in either a single or through multiple turns of interaction, respectively. In this paper, we focus on the single-turn setting, where a customer places their drink order in a single interaction. Our dataset also does not include intent classification labels, since the intent is always the same: to order drinks.

Although many datasets were released in recent years, NLU, and particularly SLU, remain scarcely covered, specially in multilingual settings. We now highlight relevant NLU/SLU resources we are aware of. \emph{SNIPS}~\cite{coucke2018snips} is a single-turn dataset covering 7 intents, such as booking a restaurant or rating a book. The \emph{ATIS} dataset~\cite{hemphill1990atis} is a popular NLU/SLU benchmark containing spoken queries related to air travel, annotated with intents (e.g., \textit{flight}, \textit{airfare}) and slots such as departure city and date. 
The \emph{SLURP} corpus~\cite{bastianelli2020slurp} offers a large-scale, multi-domain resource with real spoken utterances for virtual assistant scenarios, annotated with 18 intent types and over 50 slot categories. \emph{MASSIVE}~\cite{fitzgerald-etal-2023-massive} and \emph{Speech-MASSIVE}~\cite{lee24i_interspeech} extent a subset of SLURP to 51 text languages, and 12 speech languages, respectively.

The \emph{Fluent Speech Commands (FSC)} dataset~\cite{lugosch2019speech} consists of spoken smart home commands (e.g., ``turn on the lights in the kitchen'') annotated with triplets representing action, object, and location, effectively encoding intents and slots. 
The \emph{SmartLights} dataset~\cite{coucke2018snips,gupta2020speech} provides spoken commands for smart home device control, annotated for intent and slot-based semantics, serving as a realistic testbed for voice assistant systems. 

The \emph{Spoken Task Oriented Parsing (STOP)}~\cite{tomasello2022} dataset is a large semantically complex end-to-end spoken language dataset for end-to-end semantic parsing. It contains over 200,000 audio files from over 800 different speakers, recorded through Amazon's Mechanical Turk. 
The text utterances and semantic parses are taken from TOPv2~\cite{chen-etal-2020-low},  covering  8 different domains: alarm, event, messaging, music, navigation, reminder, timer and weather. 

Finally, The \emph{FoodOrdering dataset} \cite{a-rubino-etal-2022-cross} is  an English  NLU dataset semantically close to \emph{StarDrinks}. It is a task-oriented parsing resource focused on the food-ordering domain, 
drawing utterances and annotations from menus of five representative different venues. 
Human-generated data was crowd-sourced via Mechanical Turk,  where participants crafted natural language requests for orders serving one or multiple people based on the provided menus, 
with the resulting utterances then manually annotated into machine-executable formats.

We highlight that from these aforementioned works, only \emph{Speech-MASSIVE} present speech for the Korean language, and none covers the spoken drink ordering scenario.

\begin{figure*}
\begin{center}
\includegraphics[width=0.9\textwidth]{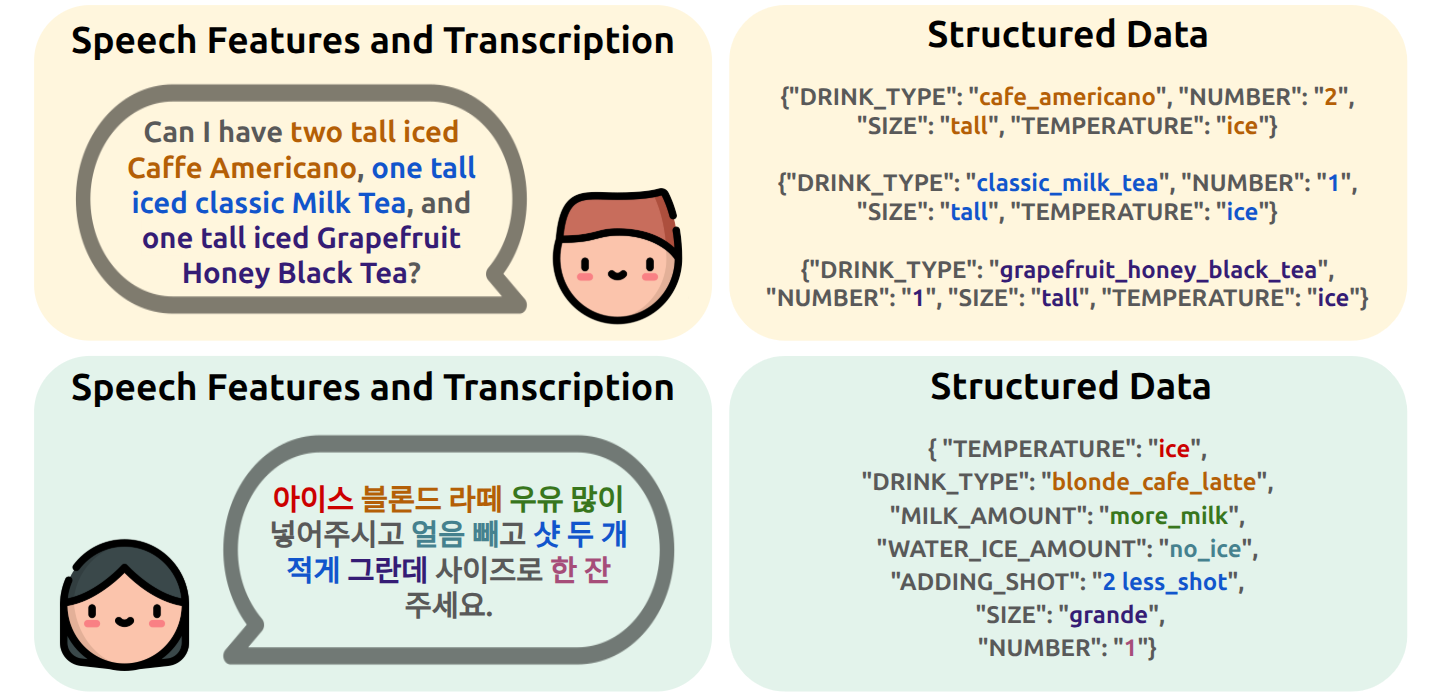}
\caption{\emph{StarDrinks} data examples from the English (top) and Korean (bottom) splits.}
\label{fig:dataexample}
\end{center}
\end{figure*}

\section{Dataset Creation}\label{sec:datacreation}

Figure~\ref{fig:datapipeline} presents the general pipeline for creating \emph{StarDrinks}. We now provide a brief overview, with further details being presented in the following sections. Starting from real drink order receipts collected from a popular coffee chain in South Korea, we extracted structured data and selected a representative subset of entries that both covered the entire menu and reflected the overall distribution of orders (Section~\ref{sec:structureddata}). 
We recruited native English and Korean speakers on the Prolific platform, assigning each participant a subset of receipts and asking them to record utterances ordering the corresponding items~(Section~\ref{sec:prolific}). From the collected speech, we generated transcripts using a state-of-the-art ASR model, followed by manual correction of both transcripts and slot annotations to produce the final output~(Section~\ref{sec:outputcleaning}). The final \emph{StarDrinks} SLU test set contains speech recordings, corresponding transcriptions, and slot annotations for each drink order.

\subsection{Structured Data Generation}\label{sec:structureddata}

We started the data generation process with 2,500 samples corresponding to real drink order receipts, which we used in order to create a large set of synthetic variations of drink order structures. % from these parsed receipts. 
Using a semantic schema that represents possible values for drink attributes~(see Table~\ref{tab:schema}), we replaced elements such as drink types, sizes, temperatures, options (e.g., milk type, syrup amount), and quantities with compatible alternatives. Low-frequency structures were prioritized to enhance diversity, resulting in a final output of up to 83,974 structures. From these, we sampled a subset that includes at least one occurrence of all attribute-value pairs in the drink orders, yielding 326 structures that are used for the Prolific annotation process.

\subsection{Data Collection}\label{sec:prolific}
\begin{figure*}
\begin{center}
\includegraphics[width=\textwidth]{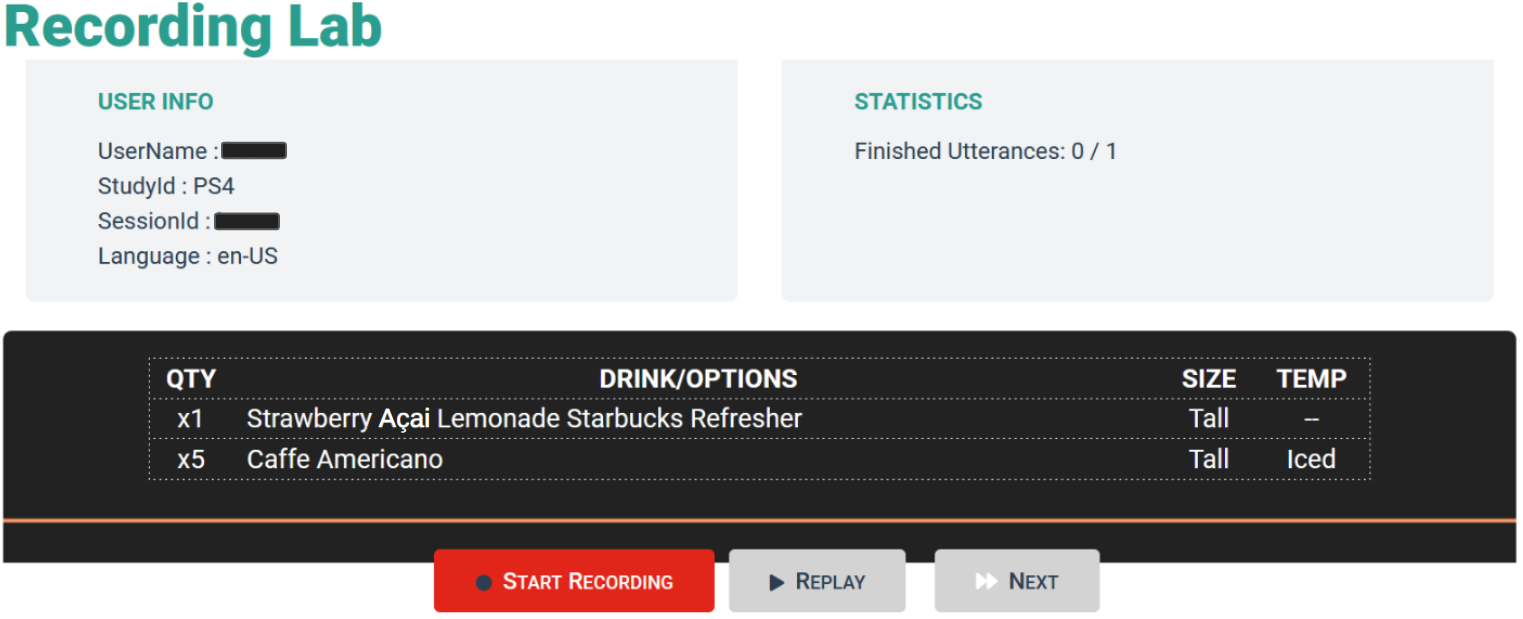}
\caption{An example from our recording session for English.}
\label{fig:prolific}
\end{center}
\end{figure*}

To gather speech recordings aligned with our collection of drink order receipts, we recruited participants via the Prolific platform.\footnote{\url{https://www.prolific.com/}} Two separate data collection campaigns were conducted, one for English and one for Korean. The participants were compensated through the standard recommendation proposed by the platform, corresponding to~£10.66 per hour. 

Before starting a session, the participants were presented with some general guidelines. Participants were asked to make a drink order in a natural manner according to the information on the screen. We also mentioned that they could use common synonyms or abbreviations for drink and option names. 

During a recording session, the recording screen presented a given receipt in English or Korean~(see Figure~\ref{fig:prolific}).  
We instructed each participant to record one utterance per drink that included all ordering options: quantity, size, customizations, and temperature. After recording, participants verified the utterance by using the “Replay” button and were allowed to re-record if they were not satisfied with the result.

For the English dataset, we recruited 32 participants located in the United Kingdom, aged between~21 and~50 years. All participants were native English speakers. Recordings were required to be made on a desktop or laptop computer equipped with a microphone.  Each participant viewed 10 receipts in a single session, and participation was restricted to one session per individual.
The demographic composition of this panel was 56.3\% male and 43.7\% female. In terms of ethnicity,~68.8\% identified as White,~15.6\% as Asian,~12.5\% as Black, and~3.1\% as mixed ethnicity. On average, participants spent 66.4 seconds recording each receipt.

For the Korean dataset, fewer eligible participants were available in the platform. Consequently, some selection criteria were relaxed. Participants were recruited from both the United Kingdom and the United States, provided they were native Korean speakers with English as a second language. Each participant could complete up to three sessions, while the age range criteria remained unchanged. 
This panel was composed of 29 participants with 31\% male and 69\% female. The average time spent recording each receipt was 51.7 seconds. 

\subsection{Output Validation}\label{sec:outputcleaning}

We collected 291 audio files for English, and 295 for Korean. We used whisper-large-v3~\cite{radford2023robust} to produce \textit{approximated} speech transcripts, since this model is very competent in English and Korean ASR. We then manually annotated the whisper output in order to validate that i)~the speech corresponds to the structured data slots; ii)~the speech corresponds to the transcript. In both cases, we corrected the slots and the transcripts when discrepancies are found, and we removed the utterance from the test set if we judged that the participants did not perform the task correctly. This annotation was performed by the authors of this paper, including fully fluent English speakers and one native Korean speaker.

We highlight that while the design of our data collection campaign forces users to produce spontaneous speech (only a receipt is shown), we did not encourage users to produce disfluent speech, which we find to be a natural consequence of collecting speech in this setting. Moreover, we did not annotate disfluent speech, as whisper is trained to ignore and correct such disfluencies. However, qualitative inspection shows that most English outputs contain either hesitation markers or repetitions.

\begin{table*}
\centering
\scriptsize
\begin{adjustbox}{width=\linewidth}
\begin{tabular}{c|c|c}
\toprule
\textbf{Slot Name}  & \textbf{Explanation}& \textbf{Possible Values}\\\midrule
\textsc{adding\_shot} & \begin{tabular}[c]{@{}c@{}}Options for adjusting the \\ number of espresso shots\end{tabular} & 1 to 6 extra\_shot, 1 to 6 less\_shot, no\_shot\\\hline
\textsc{bean} & Type of coffee bean used & blonde, decaf, half\_decaf, regular\\\hline
\textsc{chocolate\_amount}   & Intensity of chocolate flavor & light, rich\\\hline
\textsc{customs} & \begin{tabular}[c]{@{}c@{}}Drink specific \\ customizations\end{tabular} & extra\_mango\_juice, in\_ice\_cup, less\_vanilla\_cream\_base, no\_condensed\_milk, no\_tea\\\hline
\textsc{drizzle\_amount} & \begin{tabular}[c]{@{}c@{}}Adjustments to drizzle toppings \\ like caramel or chocolate\end{tabular} & \begin{tabular}[c]{@{}c@{}}extra\_chocolate\_drizzle, less\_caramel\_drizzle, less\_chocolate\_drizzle, \\ more\_caramel\_drizzle, more\_chocolate\_drizzle, regular\_chocolate\_drizzle\end{tabular}\\\hline
\textsc{drink\_type} & \begin{tabular}[c]{@{}c@{}}The specific type of beverage \\ being ordered\end{tabular} & \begin{tabular}[c]{@{}c@{}}milk\_or\_steam\_milk, coffee\_of\_the\_day, cafe\_americano, blonde\_cafe\_americano, \\ iced\_coffee, mint\_blend, youthberry, english\_breakfast, chamomile\_blend, \\ hibiscus\_blend, cafe\_latte, jeju\_organic\_green\_tea, decaffeinated\_cafe\_latte, \\ cold\_brew, blonde\_cafe\_latte, starbucks\_double\_shot, mango\_passion\_fruit\_blended,\\ espresso\_frappuccino, cold\_brew\_latte, chai\_tea\_latte, grapefruit\_honey\_black\_tea, \\ signature\_chocolate, vanilla\_cream\_cold\_brew, starbucks\_dolce\_latte, \\ decaffeinated\_starbucks\_dolce\_latte, caramel\_macchiato, white\_chocolate\_mocha, \\ blonde\_starbucks\_dolce\_latte, yuzu\_mint\_tea,blonde\_vanilla\_double\_shot\_macchiato, \\ white\_chocolate\_mocha\_frappuccino, latte\_made\_with\_jeju\_organic\_matcha, \\ classic\_milk\_tea,dolce\_cold\_brew, strawberry\_acai\_lemonade\_starbucks\_refresher,\\ pink\_drink\_with\_strawberry\_acai\_starbucks\_refresher, java\_chip\_frappuccino,\\ mango\_dragonfruit\_lemonade\_starbucks\_refresher, decaffeinated\_caramel\_macchiato, \\ chocolate\_cream\_chip\_frappuccino, cream\_frappuccino\_made\_with\_jeju\_organic\_matcha, \\ caramel\_frappuccino, strawberry\_delight\_yogurt\_blended, earl\_gray\_vanilla\_tea\_latte, \\ purple\_drink\_with\_mango\_dragonfruit\_starbucks\_refresher\end{tabular} \\\hline
\textsc{milk\_amount} & \begin{tabular}[c]{@{}c@{}}Adjustments to the amount \\ of milk\end{tabular} & less\_milk, more\_milk\\\hline
\textsc{milk\_type} & Type of milk used & low\_fat, non\_fat, oat, oat\_milk, regular\_milk, soy\\\hline
\textsc{number} & The quantity of drinks to order & 1 to 12\\\hline
\textsc{size} & Drink size options & grande, short, tall, venti\\\hline
\textsc{syrup\_amount} & Adjustments to syrup quantity & no\_syrup, no\_vanilla\_syrup\\\hline
\textsc{syrup\_type} & Syrup flavor & coffee, hazelnut, vanilla\\\hline
\textsc{temperature} & Serving temperature & hot, ice\\\hline
\textsc{water\_ice\_amount} & Adjustments to water or ice levels & extra\_ice, less\_ice, less\_water, more\_water, no\_ice \\ \hline
\textsc{whip\_cream\_amount} & \begin{tabular}[c]{@{}c@{}}Adjustments to the \\ whipped cream topping\end{tabular} & \begin{tabular}[c]{@{}c@{}}less\_espresso\_whip, less\_whip, more\_espresso\_whip, more\_whip, \\ no\_whip, regular\_espresso\_whip\end{tabular}\\
\bottomrule
\end{tabular}
\end{adjustbox}
\caption{\emph{StarDrinks} slot types, their meaning, and possible values.}
\label{tab:schema}
\end{table*}
\begin{figure*}[t]
\begin{center}
\includegraphics[width=0.9\textwidth, height=0.9\textheight, keepaspectratio]{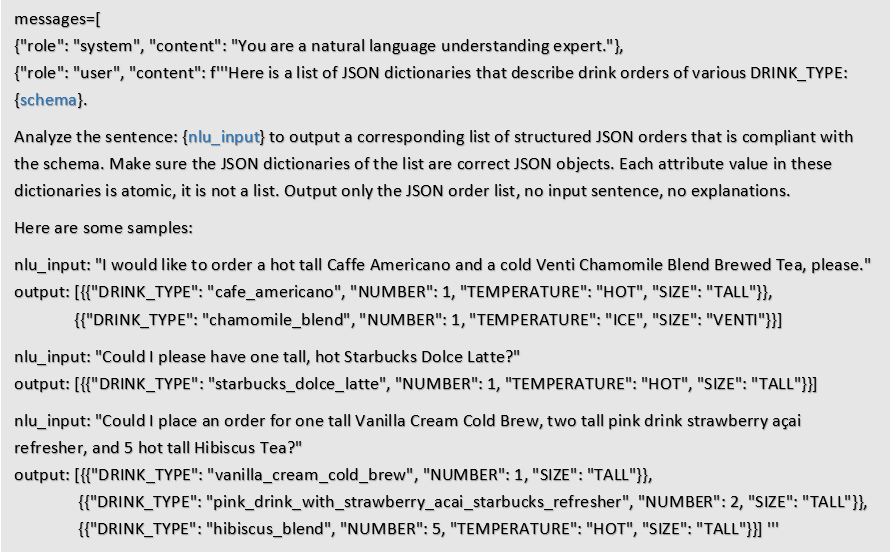}
\caption{An example of prompt for NLU (3-shots, English).}
\label{fig:prompt}
\end{center}
\end{figure*}

\section{The \emph{StarDrinks} Test Set}\label{sec:dataset}
The \emph{StarDrinks} dataset was designed to evaluate speech assistants in realistic scenarios, focusing on drink ordering in English and Korean. Each order can include up to six drinks. As a SLU dataset, it consists of speech utterance features paired with gold-standard transcriptions and structured NLU outputs. While centered on a single intent, drink ordering, the dataset features 15 distinct slots, encompassing 45 unique drink types along with their various customization options. Figure~\ref{fig:dataexample} presents examples from the dataset. Table~\ref{tab:schema} presents all existing slot values.
Statistics are reported in Table~\ref{table:stats}. 

\begin{table}
\centering
\begin{tabular}{lcc}
\toprule
 & En & Ko \\
\midrule
\# validated samples & 255 & 295 \\
\# total slots & 2,058 & 2,423 \\
\# speakers   & 32 & 29 \\\midrule
Duration & 53\,min & 45.7\,min \\
Avg. utterance length & 12.4\,s & 9.3\,s \\

\bottomrule
\end{tabular}
\caption{Statistics over \emph{StarDrinks}. }
\label{table:stats}
\end{table}

\begin{table}
\centering
\begin{tabular}{lcc}\toprule
              & \textbf{WER} & \textbf{CER} \\\midrule
English & 9.2          & 3.6          \\
Korean  & 22.9 & 7.3 \\\bottomrule   
\end{tabular}
\caption{ASR results for whisper-large-v3.}
\label{tab:asrresults}
\end{table}
\begin{table*}
\centering
\begin{adjustbox}{width=0.7\linewidth}
\begin{tabular}{ll}\toprule
\textbf{Reference} & \textbf{Whisper's Output} \\\midrule
\begin{tabular}[c]{@{}l@{}}Please can I have two \correct{cafe americano} size \\ tall and iced please?\end{tabular}                         & \begin{tabular}[c]{@{}l@{}}Please can I have two \incorrect{caffi americanas} size \\ tall and iced please?\end{tabular}                          \\
\begin{tabular}[c]{@{}l@{}}Hi, can i have one \correct{grande iced} decaf americano \\ one extra shot thank you?\end{tabular}                  & \begin{tabular}[c]{@{}l@{}}Hi, can I have one \incorrect{grand eyes} decaf americano \\ one extra shot thank you?\end{tabular}                    \\
%\begin{tabular}[c]{@{}l@{}}Can I have a grande iced cafe \correct{americano} and \\ a grande iced grapefruit honey black tea?\end{tabular}    & \begin{tabular}[c]{@{}l@{}}Can I have a grande iced cafe \incorrect{moa carne} and \\ a grande iced grapefruit honey black tea?\end{tabular}      \\
Can I get a \correct{tall strawberry yogurt}? & Can I get a \incorrect{tool to roll over your gut}? \\
\begin{tabular}[c]{@{}l@{}}Can I have one \correct{youthberry} tea size tall with ice \\ and one \correct{yuja} mint tea size tall with ice?\end{tabular} & \begin{tabular}[c]{@{}l@{}}Can I have one \incorrect{yuleberry} tea size tall with ice \\ and one \incorrect{yucca} mint tea size tall with ice?\end{tabular}\\\bottomrule
\end{tabular}
\end{adjustbox}
\caption{Some critical ASR mistakes from whisper-large-v3 on \emph{StarDrinks}.}
\label{tab:whisperexamples}
\end{table*}
\begin{table*}[ht]
\centering
\small
\begin{tabular}{l l r r r r}
\toprule
Configuration & ASR Model & \multicolumn{2}{c}{English} & \multicolumn{2}{c}{Korean} \\
\cmidrule(lr){3-4} \cmidrule(lr){5-6}
 &  & UEM (\%) & Slot F1 (\%) & UEM (\%) & Slot F1 (\%) \\
\midrule
Gold Trans. + 3-shots (NLU) & None & \textbf{87.06} & \textbf{98.04} & \textbf{89.83} & \textbf{98.76} \\
Gold Trans. + 0-shot (NLU) & None & 71.76 & 94.51 & \underline{85.76} & \underline{97.75} \\
\midrule
%GPT Transcribe / 3-shot & GPT\_4o\_transcribe & 71.76 & 93.12 & \underline{85.76} & 97.24 \\
%GPT Transcribe / 0-shot & GPT\_4o\_transcribe & 64.31 & 92.20 & 72.88 & 95.85 \\
%\midrule
ASR + 3-shots & Whisper-large-v3 & \underline{84.31} & \underline{97.37} & 84.75 & 97.45 \\
ASR + 0-shot & Whisper-large-v3 & 60.00 & 89.96 & 67.80 & 93.72 \\
\bottomrule
\end{tabular}
\caption{NLU/SLU results on the \emph{StarDrinks} English and Korean test sets with GPT-4o model.}
\label{tab:results}
\end{table*}

\section{Use Case: Drink Ordering Agent}\label{sec:usecase}

To demonstrate the utility of our test set, we evaluated a drink ordering agent built from state-of-the-art speech and text systems. We first present ASR results in Section~\ref{sec:asrresults}, illustrating the challenge of recognizing unknown named entities after training.
This is followed by slot filling results from both text and ASR output in Section~\ref{sec:slotfilling}. The agent is designed to generate NLU slots from speech input, using whisper-large-v3 for ASR and GPT-4o~\cite{achiam2023gpt} as the language model. 

\subsection{ASR results}\label{sec:asrresults}

We present the zero-shot whisper-large-v3 performance in Table~\ref{tab:asrresults}. We compute WER and CER scores using the HuggingFace evaluate library\footnote{\url{https://github.com/huggingface/evaluate}} after normalizing the input using the MMS normalization scripts from ~\citet{pratap2024scaling}.

We observe that the test splits are challenging for whisper to correctly transcribe: we reach a WER of 9.2\% for English and 22.9\% for Korean. Indeed, qualitatively, we observe that while whisper is very competent producing fluent output, since its language model did not train with this domain-specific vocabulary, it struggles to produce approximations to these new named entities. Some examples for English are given in Table~\ref{tab:whisperexamples}.

Our results highlight the ongoing challenge of adapting ASR systems to previously unseen vocabulary. In this deployment setting, one could argue that the menu is known and could be part of the system's adaptation data for fine-tuning. However, we argue that we should always expect menu changes and the introduction of new items. Therefore, we believe that a more promising open-ended solution for this problem is the research focused on context-biasing and test-time ASR adaptation~\cite{lin-etal-2024-continual,mittal-etal-2023-speech,yoon24c_interspeech}. By releasing our test set, we aim to encourage further research on these directions.

\subsection{Slot filling}\label{sec:slotfilling}

We now report on our \textit{drink ordering agent use case} NLU and SLU experiments conducted with GPT-4o as language model. For SLU, we use the ASR model from Section~\ref{sec:asrresults}. We present results for the \emph{StarDrinks} test set in both English and Korean.

In order to build our NLU component, GPT-4o was prompted to perform slot filling either on original transcriptions~(NLU) or on automatic transcriptions generated by the ASR model~(SLU). This prompt included the NLU schema with all slot types and possible slot values, and it was designed with either no examples (0-shot) or three examples (3-shot) of input and expected output (see Figure~\ref{fig:prompt}).\footnote{Additional 6 and 10 shot setups were evaluated on English NLU, but they did not yield clear improvements over the 3-shot setting.} 

We report results using two metrics: UEM and Slot F1.  
UEM (Unordered Exact Match) accuracy measures the percentage of utterances for which the entire set of predicted slot-value pairs exactly matches the reference annotation, disregarding the linear order of the pairs~\cite{rubinoCrossTOPZeroShotCrossSchema2022}. It provides a strict end-to-end measure of understanding correctness. Slot F1, instead, measures slot-level performance.

Results for English and Korean across all settings are presented in Table~\ref{tab:results}.
As expected, the highest performance is achieved with gold transcriptions, as the ASR inevitably adds noise to the input of the NLU module. For 0-shot, replacing gold transcripts by whisper-large-v3 results in an UEM accuracy reduction of 11.76 points for English, and 17.96 points for Korean. Surprisingly, this gap is much smaller in the 3-shot setting, being of only 2.75 for English, and 5.08 for Korean. This could hint to the NLU module becoming more robust to noise in this setting, potentially \textit{guessing} or correcting incorrect transcriptions given by the ASR module.

Regarding the 3-shot configuration, we find that it consistently outperformed the 0-shot setting: we observe an NLU UEM accuracy improvement of~15.3 points for English, and 4.07 for Korean. This confirms the benefit of few-shot prompting. Finally, we observe that the Korean results are generally of higher quality than the English ones (+2.77 UEM accuracy points). We believe this could be due to English being a language that allows for a more flexible expression of the items in an order, making this test set slightly more challenging.

Although the best UEM accuracy scores appear reasonably high, they remain relatively low for practical user-facing applications, where near-perfect understanding is required. In those settings, Substantial amounts of adaptation data, whether collected in-domain or synthetically generated, would likely be required to achieve significant improvements in the agent's performance.
\section{Conclusion}\label{sec:conclusion}

In this paper we presented \emph{StarDrinks}, a test set consisting of spontaneously uttered drink orders, their transcripts and NLU slot values. It provides data for speech-to-slots SLU, text-to-slots NLU and speech-to-transcription ASR evaluation in both English and Korean. 

To showcase its usefulness, we presented a coffee ordering agent baseline by plugging whisper-large-v3 to GPT-4o. We observe that the ASR module struggles to adapt to unknown named entities, highlighting the necessity of research on test-time adaptation approaches. Regarding NLU/SLU results, we observe that, while the reported UEM accuracy can get as high as~87.06\% for English and~89.83\% for Korean for a given setting, this performance is still short of the near-perfect requirements for a deployed system. 

We hope our test set will provide a realistic and challenging adaptation setting for NLU and SLU models, supporting
the development of more robust, context-aware systems capable of handling real-world interactions. The dataset is available for download at \url{https://europe.naverlabs.com/stardrinks}.

{
    \small
    \bibliographystyle{ieeenat_fullname}
    \bibliography{main}
}

\end{document}